\documentclass{article}
 \PassOptionsToPackage{numbers, compress}{natbib}

\usepackage[final]{neurips_2019}

\usepackage[utf8]{inputenc} 
\usepackage[T1]{fontenc}    
\usepackage{hyperref}       
\usepackage{booktabs}       
\usepackage{amsfonts}       
\usepackage{nicefrac}       
\usepackage{microtype}      
\usepackage{authblk}
\usepackage{times}
\usepackage{soul}
\usepackage{url}
\usepackage[small]{caption}
\usepackage{graphicx}
\usepackage{subfig}
\usepackage{amsmath}
\usepackage{booktabs}
\usepackage{algorithm}
\usepackage{algorithmic}
\usepackage{amsthm}
\usepackage{amssymb}
\usepackage[american]{babel}
\usepackage{footnote}
{

}
\usepackage{mathtools}

\title{Trained Rank Pruning for Efficient Deep Neural Networks}

\author[1]{Yuhui~Xu}
\author[1]{Yuxi~Li}
\author[2]{Shuai~Zhang}
\author[3]{Wei Wen}
\author[2]{Botao~Wang}
\author[1]{Wenrui~Dai}
\author[2]{Yingyong~Qi}
\author[3]{Yiran Chen}
\author[1]{Weiyao~Lin}
\author[1]{Hongkai~Xiong}
\affil[1]{Shanghai Jiao Tong University, 
	Email: \{yuhuixu, lyxok1, daiwenrui, wylin, xionghongkai\}@sjtu.edu.cn}
\affil[2]{Qualcomm AI Research,
	Email: \{shuazhan, botaow, yingyong\}@qti.qualcomm.com}
\affil[3]{Duke University,
	Email: \{wei.wen, yiran.chen\}@duke.edu}

\begin{document}

\maketitle

\begin{abstract}
To accelerate DNNs inference, low-rank approximation has been widely adopted because of its solid theoretical rationale and efficient implementations. Several previous works attempted to directly approximate a pre-trained model by low-rank decomposition; however, small approximation errors in parameters can ripple over a large prediction loss. Apparently, it is not optimal to separate low-rank approximation from training. Unlike previous works, this paper integrates low rank approximation and regularization into the training process. We propose Trained Rank Pruning (TRP), which alternates between low rank approximation and training. TRP maintains the capacity of the original network while imposing low-rank constraints during training. A nuclear regularization optimized by stochastic sub-gradient descent is utilized to further promote low rank in TRP. Networks trained with TRP has a low-rank structure in nature, and is approximated with negligible performance loss, thus eliminating fine-tuning after low rank approximation. The proposed method is comprehensively evaluated on CIFAR-10 and ImageNet, outperforming previous compression counterparts using low rank approximation. Our code is available at: \url{https://github.com/yuhuixu1993/Trained-Rank-Pruning}.
\end{abstract}

\section{Introduction}

Deep Neural Networks (DNNs) have shown remarkable success in many computer vision tasks such as image classification \cite{He2016DeepRL}, object detection \cite{Ren2015FasterRT} and semantic segmentation \cite{Chen2018DeepLabSI}. 
Despite the high performance in large DNNs powered by cutting-edge parallel computing hardware, most of state-of-the-art network architectures are not suitable for resource restricted usage such as usages on always-on devices, battery-powered low-end devices, due to the limitations on computational capacity, memory and power. 

To address this problem, low-rank decomposition methods \cite{Denton2014Exploiting,jaderberg2014speeding,Guo2018Network,Wen2017Coordinating,Alvarez2017Compression} have been proposed to minimize the channel-wise and spatial redundancy by decomposing the original network into a compact one with low-rank layers. Different from precedent works, this paper proposes a novel approach to design low-rank networks.

Low-rank networks can be trained directly from scratch. However, it is difficult to obtain satisfactory results for several reasons.
(1)~\textit{Low capacity:}
Compared with the original full rank network, the capacity of a low-rank network is limited, which causes difficulties in optimizing its performances. 
(2)~\textit{Deep structure:}
Low-rank decomposition typically doubles the number of layers in a network. The additional layers make numerical optimization much more challenging because of exploding and/or vanishing gradients.
(3)~\textit{Rank selection:}
The rank of decomposed network is often  chosen as a hyperparameter based on pre-trained networks; which may not be the optimal rank for the network trained from scratch.

Alternatively, several previous works \cite{zhang2016accelerating,Guo2018Network,jaderberg2014speeding} attempted to decompose pre-trained models in order to get initial low-rank networks. However, the heuristically imposed low-rank could incur huge accuracy loss and network retraining is required to recover the performance of the original network as much as possible. Some attempts were made to use sparsity regularization \cite{Wen2017Coordinating,chen2015compressing} to constrain the network into a low-rank space. Though sparsity regularization reduces the error incurred by decomposition to some extent, performance still degrades rapidly when compression rate increases.

\begin{figure*}[h]
	\centering
	\includegraphics[width=5.5in]{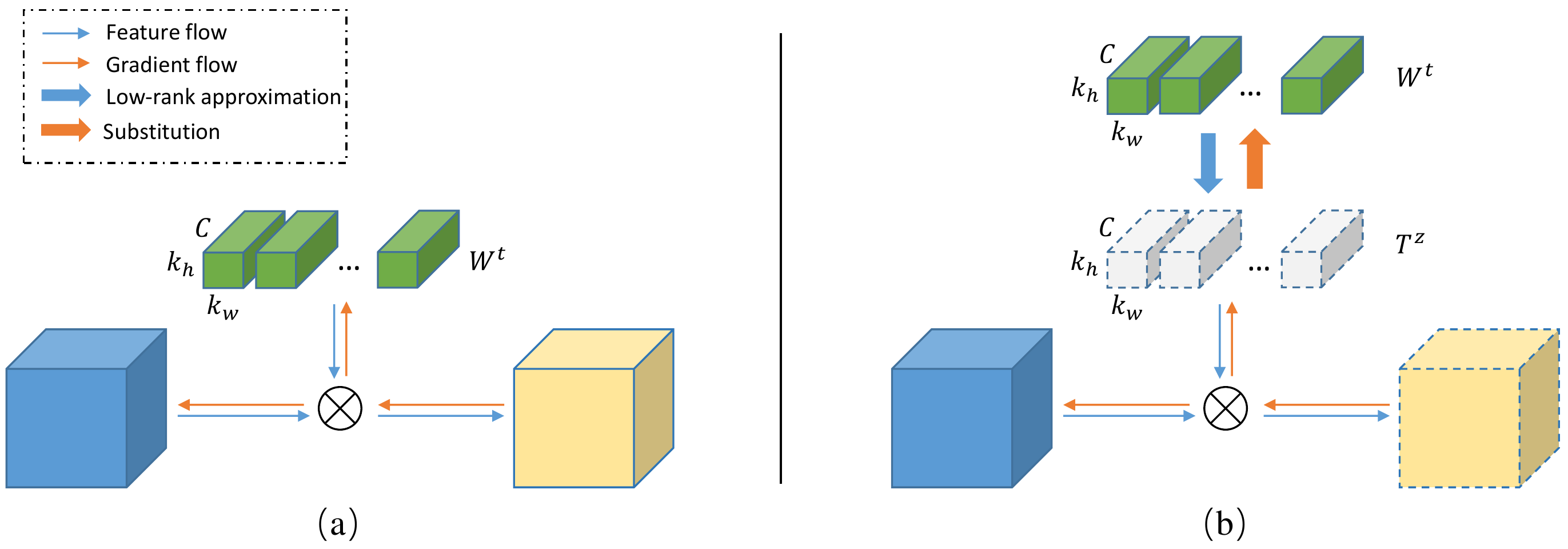}
	\caption{The training of TRP consists of two parts as illustrated in (a) and (b). (a) One normal iteration with forward-backward broadcast and weight update. (b) One training iteration inserted by rank pruning, where the low-rank approximation is first applied on current filters before convolution. During backward propagation, the gradients are directly added on low-rank filters and the original weights are substituted by updated low-rank filters. (b) is applied once every $m$ iterations (\textit{i.e.} when gradient update iteration $t=zm, z=0, 1, 2, \cdots$), otherwise (a) is applied.}\label{fig.1}
\end{figure*}

In this paper, we propose a new method, namely Trained Rank Pruning (TRP), for training low-rank networks. We embed the low-rank decomposition into the training process by gradually pushing the weight distribution of a well functioning network into a low-rank form, where all parameters of the original network are kept and optimized to maintain its capacity. We also propose a stochastic sub-gradient descent optimized nuclear regularization that further constrains the weights in a low-rank space to boost the TRP.
 The proposed solution is illustrated in Fig.~\ref{fig.1}.

Overall, our contributions are summarized below.
\begin{enumerate}
    \setlength\itemsep{-0em}
    \item A new training method called TRP is presented by explicitly embedding the low-rank decomposition into the network training;
    \item A nuclear regularization is optimized by stochastic sub-gradient descent to boost the performance of the TRP;
    \item Improving inference acceleration and reducing approximation accuracy loss in both channel-wise and spatial-wise decomposition methods.  
\end{enumerate}

\section{Methodology}
\subsection{Preliminaries}
Formally, the convolution filters in a layer can be denoted by a tensor $W \in \mathbb{R}^{n \times c \times k_w \times k_h}$, where $ n $ and $ c $ are the number of filters and input channels, $ k_h $ and $ k_w $ are the height and width of the filters. An input of the convolution layer $ F_i \in  \mathbb{R}^{c \times x \times y}$ generates an output as $ F_o = W * F_i $. Channel-wise correlation \cite{zhang2016accelerating} and spatial-wise correlation \cite{jaderberg2014speeding} are explored to approximate convolution filters in a low-rank space. In this paper, we focus on these two decomposition schemes. 
However, unlike the previous works, we propose a new training scheme TRP to obtain a low-rank network without re-training after decomposition.

\subsection{Trained Rank Pruning}
We propose a simple yet effective training scheme called Trained Rank Pruning (TRP) in a periodic fashion:

\begin{equation}\label{equ.2}
\begin{split}
W^{t+1}&=\left \{ 
\begin{array}{c}
     W^t - \alpha \triangledown f(W^t)  \quad t \% m \neq 0 \\
     T^z - \alpha \triangledown f(T^z)  \quad t \% m = 0 
\end{array} \right. \\
T^{z}&=\mathcal{D}(W^{t}), \quad z = t / m
\end{split}
\end{equation}
where $\mathcal{D}({\cdot})$ is a  low-rank  tensor approximation operator, $\alpha$ is the learning rate, $t$ indexes the iteration and $z$ is the iteration of the operator $\mathcal{D}$ , with $m$ being the period for the low-rank approximation.


We apply low-rank approximation every $m$ SGD iterations. This saves training time to a large extent. As illustrated in Fig.~\ref{fig.1}, for every $m$ iterations, we perform low-rank approximation on the original filters, while gradients are updated on the resultant low-rank form. Otherwise, the network is updated via the normal SGD. 
Our training scheme could be combined with arbitrary low-rank operators. In the proposed work, we choose the low-rank techniques proposed in \cite{jaderberg2014speeding} and \cite{zhang2016accelerating}, both of which transform the 4-dimensional filters into 2D matrix and then apply the truncated singular value decomposition (TSVD). The SVD of matrix ${W}^t$ can be written as:
\begin{equation}\label{equ.3}
W^t=\sum_{i=1}^{rank(W^t)}\sigma_i\cdot U_i\cdot (V_i)^T,
\end{equation}
where $\sigma_i$ is the singular value of $W^t$ with $\sigma_1\geq \sigma_2 \geq \cdots \geq \sigma_{rank(W^t)}$, and $U_i$ and $V_i$ are the singular vectors. The parameterized TSVD ($W^t;e$) is to find the smallest integer $k$ such that
\begin{equation}\label{equ.4}
\sum_{j=k+1}^{rank(W^t)}(\sigma_j)^2 \leq \quad e \sum_{i=1}^{rank(W^t)}(\sigma_i)^2,
\end{equation}
where $e\in(0 , 1)$ is a pre-defined hyper-parameter of the energy-preserving ratio. After truncating the last $n-k$ singular values, we transform the low-rank 2D matrix  back to 4D tensor.

\subsection{Nuclear Norm Regularization}
Nuclear norm is widely used in matrix completion problems. Recently, it is introduced to constrain the network into low-rank space during the training process \cite{Alvarez2017Compression}.
\begin{equation}\label{equ.5}
\min \left\{  f\left(x; w\right)+\lambda \sum_{l=1}^L||W_l||_* \right\}
\end{equation}
where $f(\cdot)$ is the objective loss function, nuclear norm $||W_l||_*$ is defined as $||W_l||_*=\sum_{i=1}^{rank(W_l)}\sigma_l^i$, with $\sigma_l^i$ the singular values of $W_l$. $\lambda$ is a hyper-parameter setting the influence of the nuclear norm. In this paper, we utilize stochastic sub-gradient descent \cite{Avron2012EfficientAP} to optimize nuclear norm regularization in the training process. Let $W=U\Sigma V^T$ be the SVD of $W$ and let $U_{tru}, V_{tru}$ be $U, V$ truncated to the first $rank(W)$ columns or rows, then $U_{tru}V_{tru}^T$ is the sub-gradient of $||W||_*$ \cite{watson1992characterization}. Thus, the sub-gradient of  Eq.~(\ref{equ.5}) in a layer is 
\begin{equation}\label{equ.6}
\triangledown f+\lambda U_{tru}V_{tru}^T.
\end{equation}

The nuclear norm and loss function are optimized simultaneously during the training of the networks and can further be combined with the proposed TRP.

\begin{table}[htb]
\small

\subfloat{
\begin{minipage}{0.58\textwidth}
\centering
\begin{tabular}{|l|c|c|c|}
\hline
Model ("R-" indicates ResNet-.) & Top 1 ($\%$) & Speed up\\
\hline\hline
R-20 (baseline)&91.74&1.00$\times$\\
\hline
R-20 (TRP1)&90.12&1.97$\times$\\
R-20 (TRP1+Nu)&\textbf{90.50}&\textbf{2.17$\times$}\\
R-20 (\cite{zhang2016accelerating})&88.13&1.41$\times$\\
\hline
R-20 (TRP2)&90.13&2.66$\times$\\
R-20 (TRP2+Nu)&\textbf{90.62}&\textbf{2.84$\times$}\\
R-20 (\cite{jaderberg2014speeding})&89.49&1.66$\times$\\
\hline
\hline
R-56 (baseline)&93.14&1.00$\times$\\
\hline
R-56 (TRP1)&\textbf{92.77}&2.31$\times$\\
R-56 (TRP1+Nu)&91.85&\textbf{4.48$\times$}\\
R-56 (\cite{zhang2016accelerating})&91.56&2.10$\times$\\
\hline
R-56 (TRP2)&\textbf{92.63}&2.43$\times$\\
R-56 (TRP2+Nu)&91.62&\textbf{4.51$\times$}\\
R-56 (\cite{jaderberg2014speeding})&91.59&2.10$\times$\\
\hline
R-56 \cite{He_2017_ICCV}&91.80&2.00$\times$\\
R-56 \cite{li2016pruning}&91.60&2.00$\times$\\
\hline
\end{tabular}
\caption{Experiment results on CIFAR-10. }\label{tab1}
\vspace{-0.4cm} 
\end{minipage}
}
\hfill
\subfloat{
\begin{minipage}{0.45\textwidth}
\centering
\begin{tabular}{|l|c|c|}
\hline
Method &Top1($\%$)& Speed up\\
\hline\hline
Baseline&69.10&1.00$\times$\\
\hline
TRP1 &\textbf{65.46} &1.81$\times$\\
TRP1+Nu&65.39&\textbf{2.23$\times$}\\
\cite{zhang2016accelerating}&63.1&1.41$\times$\\
\hline
TRP2 &\textbf{65.51}&$2.60\times$\\
TRP2+Nu&65.34&\textbf{3.18}$\times$\\
\cite{jaderberg2014speeding}&62.80&2.00$\times$\\
\hline
\end{tabular}
\caption{Results of ResNet-18 on ImageNet.}\label{tab2}

\begin{tabular}{|l|c|c|}
\hline
Method &Top1($\%$)& Speed up\\
\hline\hline
Baseline&75.90&1.00$\times$\\
\hline
TRP1+Nu&72.69&\textbf{2.30}$\times$\\
TRP1+Nu&\textbf{74.06}&1.80$\times$\\
\cite{zhang2016accelerating}&71.80&1.50$\times$\\
\cite{Luo2017ThiNetAF}&72.04&1.58\\
\cite{luo2018thinet}&72.03&2.26\\
\hline
\end{tabular}
\caption{Results of ResNet-50 on ImageNet.}\label{tab3}
\vspace{-0.4cm} 
\end{minipage}
}

\end{table}

\section{Experiments}\label{sec:exp}

\subsection{Implementation Details}
We evaluate the performance of TRP scheme on two common datasets, CIFAR-10 \cite{AlexCifar10} and ImageNet \cite{Deng2009ImageNetAL}. 
We implement our TRP scheme with NVIDIA 1080 Ti GPUs. For training on CIFAR-10, we start with base learning rate of $0.1$ to train 164 epochs and degrade the value by a factor of $10$ at the $82$-th and $122$-th epoch. For ImageNet, we directly finetune the model with TRP scheme from the pre-trained baseline with learning rate $0.0001$ for 10 epochs. For both of the datasets, we adopt SGD solver to update weight and set the weight decay value as $10^{-4}$ and momentum value as $0.9$.
\footnotetext[1]{the implementation of \cite{Guo2018Network}} 

\subsection{Results on CIFAR-10}
 As shown in Table~\ref{tab1}, for both spatial-wise (TRP1) and channel-wise (TRP2) decomposition, the proposed TRP outperforms basic methods~\cite{zhang2016accelerating,jaderberg2014speeding} on ResNet-20 and ResNet-56. Results become even better when nuclear regularization is used. For example, in the channel-wise decomposition (TRP2) of ResNet-56, results of TRP combined with nuclear regularization can even achieve $2\times$ speed up rate than \cite{zhang2016accelerating} with same accuracy drop. Our method also outperforms filter pruning~\cite{li2016pruning} and channel pruning~\cite{He_2017_ICCV}. For example, the channel decomposed TRP trained ResNet-56 can achieve $92.77\%$ accuracy with $2.31\times$ acceleration, while \cite{He_2017_ICCV} is $91.80\%$ and \cite{li2016pruning} is $91.60\%$. With the help of nuclear regularization, our methods can obtain $2$ times of the acceleration rate of \cite{He_2017_ICCV} and \cite{li2016pruning} with higher accuracy.


%

\subsection{Results on ImageNet}

The results on ImageNet are shown in Table~\ref{tab2} and Table~\ref{tab3}. For ResNet-18, our method outperforms the basic methods \cite{zhang2016accelerating,jaderberg2014speeding}. For example, in the channel-wise decomposition, TRP obtains 1.81$\times$ speed up rate with 86.48\% Top5 accuracy on ImageNet which outperforms both the data-driven \cite{zhang2016accelerating}\textsuperscript{1} and data independent \cite{zhang2016accelerating} methods by a large margin. Nuclear regularization can increase the speed up rates with the same accuracy.

For ResNet-50, to better validate the effectiveness of our method, we also compare the proposed TRP with \cite{He_2017_ICCV} and \cite{Luo2017ThiNetAF}. With $1.80\times$ speed up, our decomposed ResNet-50 can obtain $73.97\%$ Top1 and $91.98\%$ Top5 accuracy which is much higher than \cite{Luo2017ThiNetAF}. The TRP achieves $2.23\times$ acceleration which is higher than \cite{He_2017_ICCV} with the same Top5 degrade.

\section{Conclusion}

In this paper, we propose a new scheme Trained Rank Pruning (TRP) for training low-rank networks. It leverages capacity and structure of the original network by embedding the low-rank approximation in the training process. Furthermore, we propose stochastic sub-gradient descent optimized nuclear norm regularization to boost the TRP. The proposed TRP can be incorporated with any low-rank decomposition method. On CIFAR-10 and ImageNet datasets, we have shown that our methods can outperform basic methods both in channel-wise decmposition and spatial-wise decomposition.
\clearpage
\bibliographystyle{abbrv}
\bibliography{ijcai19}

\end{document}